\documentclass{article}

\usepackage{microtype}
\usepackage{graphicx}
\usepackage{subfigure}
\usepackage{booktabs} 

\usepackage{hyperref}

\usepackage[accepted]{icml2025}

\usepackage{amsmath}
\usepackage{amssymb}
\usepackage{mathtools}
\usepackage{amsthm}
\usepackage{multirow}

\usepackage[capitalize,noabbrev]{cleveref}

\theoremstyle{plain}

\theoremstyle{definition}

\theoremstyle{remark}

\usepackage[textsize=tiny]{todonotes}

\icmltitlerunning{ADHMR: Aligning Diffusion-based Human Mesh Recovery via Direct Preference Optimization}

\begin{document}

\twocolumn[
\icmltitle{ADHMR: Aligning Diffusion-based Human Mesh Recovery \\ via Direct Preference Optimization}




\begin{icmlauthorlist}
\icmlauthor{Wenhao Shen}{ntu}
\icmlauthor{Wanqi Yin}{sensetime}
\icmlauthor{Xiaofeng Yang}{ntu}
\icmlauthor{Cheng Chen}{ntu}
\icmlauthor{Chaoyue Song}{ntu}
\icmlauthor{Zhongang Cai}{sensetime}
\icmlauthor{Lei Yang}{sensetime}
\icmlauthor{Hao Wang}{hkust}
\icmlauthor{Guosheng Lin}{ntu}
\end{icmlauthorlist}

\icmlaffiliation{ntu}{Nanyang Technological University}
\icmlaffiliation{sensetime}{SenseTime Research}
\icmlaffiliation{hkust}{The Hong Kong University of Science and Technology (Guangzhou)}

\icmlcorrespondingauthor{Guosheng Lin}{gslin@ntu.edu.sg}
\icmlcorrespondingauthor{Hao Wang}{haowang@hkust-gz.edu.cn}

\icmlkeywords{Human mesh recovery, Direct preference optimization}

\vskip 0.3in
]



\printAffiliationsAndNotice{}  

\begin{abstract}
Human mesh recovery (HMR) from a single image is inherently ill-posed due to depth ambiguity and occlusions. Probabilistic methods have tried to solve this by generating numerous plausible 3D human mesh predictions, but they often exhibit misalignment with 2D image observations and weak robustness to in-the-wild images. To address these issues, we propose ADHMR, a framework that \textbf{A}ligns a \textbf{D}iffusion-based \textbf{HMR} model in a preference optimization manner. First, we train a human mesh prediction assessment model, HMR-Scorer, capable of evaluating predictions even for in-the-wild images without 3D annotations. We then use HMR-Scorer to create a preference dataset, where each input image has a pair of winner and loser mesh predictions. This dataset is used to finetune the base model using direct preference optimization. Moreover, HMR-Scorer also helps improve existing HMR models by data cleaning, even with fewer training samples. Extensive experiments show that ADHMR outperforms current state-of-the-art methods. Code is available at: \textit{\href{https://github.com/shenwenhao01/ADHMR}{\textcolor{magenta}{https://github.com/shenwenhao01/ADHMR}}}.
\end{abstract}

\section{Introduction}
\label{sec:intro}

Human mesh recovery (HMR) is a fundamental challenge in computer vision, focused on estimating the 3D human shape and pose from a single RGB image. 
HMR enables various downstream applications, including clothed human reconstruction~\cite{multinb, hong2024free, yao2025unify3d}, virtual try-on, AR/VR content creation~\cite{xu2024high, yang2024attrihuman} and etc.

Prevailing approaches usually adopt a deterministic style, generating a single prediction for each image~\cite{cai2024smplerx, goel2023hmr2.0, li2023hybrikx, moon2022h4w}.
However, this task faces inherent uncertainty when lifting 2D observations to 3D models, due to depth ambiguity and occlusion.
Accordingly, the community is now shifting to probabilistic methods. 
Probabilistic methods tackle the uncertainty by generating multiple plausible human mesh predictions for each image~\cite{kolotouros2021prohmr, sengupta2023humaniflow}.
For instance, recent approaches~\cite{foo2023hmdiff,cho2023diffhmr} frame this task as a denoising diffusion process.
However, these probabilistic approaches suffer from limited emphasis on obtaining accurate estimates.

Specifically, the current state-of-the-art probabilistic method ScoreHypo~\cite{xu2024scorehypo} tackles this by designing an additional network for test-time prediction selection after the diffusion-based prediction model.
However, we observe that ScoreHypo still exhibits the following shortcomings: 
(1) misalignment between 3D mesh predictions and 2D image cues, and (2) poor performance on in-the-wild images. 
This is primarily because end-to-end diffusion models predicting from pure noise typically avoid 3D reprojection loss, as early denoising steps yield unrealistic poses, making such loss ineffective~\cite{huang2024closeint}.
Instead, the diffusion loss focuses on generating the target human mesh distribution rather than precisely aligning joints. While this produces plausible poses, it may neglect the alignment between the 3D mesh and the image. 
Moreover, existing datasets often use optimization-based HMR methods to generate pseudo 3D annotations for in-the-wild images, which inevitably contain some inaccurate or noisy data.

To address the above challenges, we introduce Aligned Diffusion-based Human Mesh Recovery (ADHMR).
The key insight is to distill the knowledge of a powerful scorer into the 3D human mesh predictor in a preference optimization manner. 
Technically, we begin by training the HMR-Scorer, essentially a reward model that assigns a quality score to quantify the human mesh prediction quality.
HMR-Scorer gives higher scores to predictions better aligned with the input image (``winners'') and lower scores to those poorly aligned (``losers'').
In order to increase the sensitivity of HMR-Scorer to nuances in the image cues, we extract multi-scale image features as conditions, which provide global and local pixel-aligned features sampled by reprojected human keypoints, enabling HMR-Scorer to identify misalignment between the predicted mesh and 2D image cues.

Then we draw on the concept of direct preference optimization (DPO) for diffusion models~\cite{wallace2024diffusiondpo} to optimize a diffusion-based HMR base model.
Traditional joint-wise or pixel-wise losses could overfit noisy labels in real-world data. Besides, a trade-off between 2D reprojection fidelity and 3D accuracy exists due to imprecise camera estimation~\cite{dwivedi2024tokenhmr}.
In contrast, DPO focuses on the relative prediction quality, being more robust to imperfect data.
However, DPO requires an annotated preference dataset, which is costly to obtain~\cite{rafailov2024dpo}.
To this end, we employ HMR-Scorer to evaluate and rank the predictions generated by the base model, resulting in a preference dataset composed of $\left<\text{winner}, \text{loser}\right>$ prediction pairs. 
Guided by this synthetic dataset, ADHMR refines the HMR base model towards producing human pose predictions that are both more plausible and more closely aligned with 2D image cues.
Moreover, ADHMR improves its robustness by finetuning on in-the-wild images without the need for pseudo labels.

Notably, HMR-Scorer can also be leveraged to improve the performance of state-of-the-art HMR models through data cleaning. 
Many models~\cite{yin2025smplest, pang2024robosmplx, sun2024aios} incorporate in-the-wild datasets for training to enhance their generalizability. However, as mentioned earlier, the 3D pseudo-labels in these datasets are often unreliable.
Prior work relies on expensive manual curation~\cite{lassner2017unite} or rigid reprojection-error filtering~\cite{kolotouros2019spin} to combat these issues.
Instead, we propose to conduct a fully automated data cleaning process to build higher-quality training datasets. We sort the pseudo-labeled images in a dataset based on their scores given by HMR-Scorer and only retain samples with scores above a certain threshold. 
By filtering out poorly annotated data, we reduce the influence of noisy annotations and boost model performance.

Comprehensive experimental results demonstrate the effectiveness of our approach. 
The main contributions are summarized below:
\begin{itemize} 
    \item We propose ADHMR, a novel framework for improving existing diffusion-based HMR models by adapting human preference optimization methods to an unlabeled setting, thus outperforming existing state-of-the-art probabilistic HMR methods.
    \item We introduce HMR-Scorer, a robust reward model that effectively quantifies the alignment between human mesh predictions and corresponding input images. 
    \item We show that using HMR-Scorer for data cleaning boosts the performance of state-of-the-art HMR models, even when trained on fewer data.
\end{itemize}
\section{Related work}
\label{sec:related}

\subsection{Human Mesh Recovery from a Single Image}
Current HMR approaches can be broadly categorized into two paradigms: deterministic and probabilistic.
Deterministic approaches\cite{goel2023hmr2.0, cai2024smplerx, moon2022h4w, hmradapter, yin2024whac} produce a single estimate for each input. 
However, due to intrinsic reconstruction ambiguities, probabilistic approaches focus on generating multiple plausible hypotheses or capturing probabilistic distributions.
ProHMR~\cite{kolotouros2021prohmr} leverages a conditional normalizing flow to model a conditional probability distribution.
Fang et al.~\cite{fang2023posterior} propose learning probability distributions over human joint rotations by utilizing a learned analytical posterior probability.
EgoHMR~\cite{zhang2023egohmr} proposes a 3D scene-conditioned diffusion approach for reconstructing human meshes from egocentric views.
ScoreHypo~\cite{xu2024scorehypo} uses a diffusion-based generator to produce a diverse set of plausible estimates, and a separate network is employed to choose from these estimates.
Despite their effectiveness, they often require generating numerous candidate poses for selection or averaging. 
In contrast, we employ direct preference optimization to improve the performance of the prediction model directly.

\begin{figure*}
    \centering
    \includegraphics[width=1\linewidth]{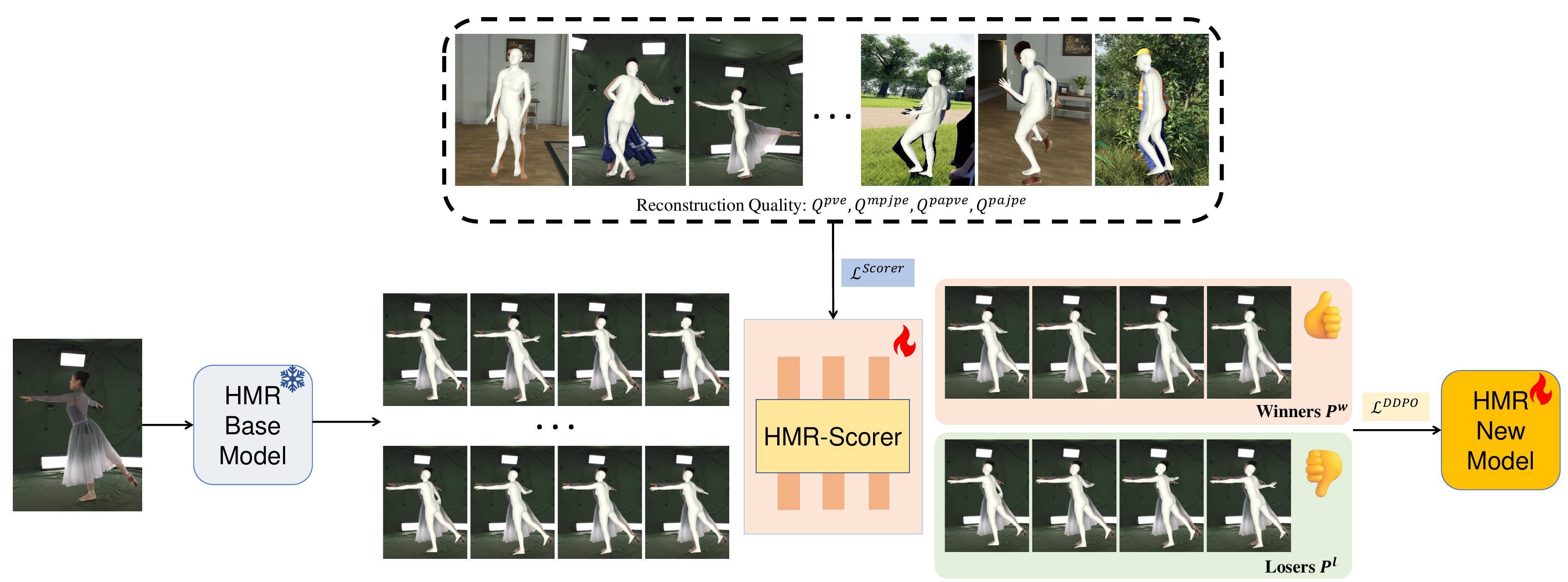}
    \vspace{-20pt}
    \caption{
    Overview of ADHMR.
    We aim to finetune a probabilistic HMR base model that generates multiple human mesh predictions conditioned on the input image.
    We first train the HMR-Scorer that assesses the reconstruction quality given an image and corresponding human mesh predictions.
    The reconstruction quality annotations $Q^*$ are computed using standard HMR metrics, including PVE $Q^{pve}$, MPJPE $Q^{mpjpe}$, PA-MPJPE $Q^{pajpe}$, and PA-PVE $Q^{papve}$.
    Next, we construct a synthetic human preference dataset, where each sample is a $\left<\text{winner}, \text{loser}\right>$ prediction pair rated by the HMR-Scorer.
    Finally, ADHMR uses this synthetic human preference dataset to finetune the base model to preferentially generate predictions that are more plausible and better aligned with the image cues.
    \vspace{-10pt}
    }
\label{fig:overview}
\end{figure*}

\subsection{Human Preference Optimization}

The initial efforts to learn from human preferences originated in training agents~\cite{christiano2017rlhf, ibarz2018reward}, later expanding to incorporate human feedback (RLHF) for enhancing tasks like translation~\cite{kreutzer2018reliability} and summarization~\cite{stiennon2020learning, ziegler2019fine}.
These methods first train a reward model to align with human preferences and then finetune a language model to maximize this reward using reinforcement learning techniques such as PPO~\cite{schulman2017ppo}.
Several solutions have been proposed to simplify this complex pipeline: 
HIVE ~\cite{zhang2024hive} uses offline reinforcement learning to align instruction editing.
Direct Preference Optimization (DPO)~\cite{rafailov2024dpo} directly optimizes the model using a supervised classification objective on preference data.
This approach is now being increasingly adopted across other domains.
For instance, ImageReward~\cite{xu2024imagereward} and Lee et al. ~\cite{lee2023aligning} apply RLHF to text-to-image synthesis models; DreamReward~\cite{ye2024dreamreward} and CADCrafter~\cite{chen2025cadcrafter} use RLHF for text-to-3D generation. 
Diffusion-DPO~\cite{wallace2024diffusiondpo} adapts the DPO objective for Diffusion Models, improving the performance of models like Stable Diffusion for enhanced visual appeal and textual coherence.
Our method is inspired by Diffusion-DPO but differs in its implementation. Rather than depending on curated manually labeled human feedback datasets, we devise a method to automatically generate a human preference dataset using HMR-Scorer, offering greater flexibility for our scenario.

\section{Preliminary}

\subsection{HMR Evaluation Metrics}
\label{subsec:hmrmetrcs}
We use four standard metrics for HMR: Mean Per Vertex Position Error (PVE) and Mean Per Joint Position Error (MPJPE), along with their Procrustes-aligned variants (PA-PVE and PA-MPJPE). All metrics compute the average distance (in mm) between predicted and ground-truth positions, with the pelvis joint aligned as the reference point. We apply the joint regressor of SMPL(-X) to the predicted mesh to obtain 3D joint coordinates.

\subsection{Diffusion-based HMR Base Model}
\label{sec:scorehypo}
Modeling HMR as a reverse diffusion process by noisy samples $\{\mathbf{x}_{t}\}_{t=0}^T$, the base model HypoNet~\cite{xu2024scorehypo} $\boldsymbol{\epsilon}_\text{ref}$ is a denoiser that progressively denoises random pose noise based on the input image $I$ to reconstruct the human mesh. $T$ is the total number of timesteps. This process is formulated as:
\begin{equation}
    p_\theta\left(\mathbf{x}_{t-1} \mid \mathbf{x}_t\right)=\mathcal{N}\left(\mathbf{x}_{t-1} ; \boldsymbol{\epsilon}_\text{ref}\left(\mathbf{x}_t, t, I\right)\right),
\end{equation}
where $p_\theta$ is the posterior mean of the forward process.

Specifically, the base model follows ~\cite{li2021hybrik} by breaking down the SMPL~\cite{loper2015smpl} pose parameters into two components: swing, derived from 3D joint positions, and twist, representing rotational details for each body part. These two elements are combined into a single data sample, which is then processed through a forward diffusion step to gradually add noise.
The noisy samples are mapped to a high-dimensional feature space using a multilayer perceptron. To guide the denoising process, the model incorporates image features extracted through a convolutional neural network backbone. 
These preprocessed image features are concatenated and fed into a transformer-encoder~\cite{vaswani2017attention} based network. The transformer integrates global image context through a cross-attention mechanism, aligning the denoising process with the input image. 
Finally, the network reconstructs the human pose by removing the added noise. The human shape parameters are estimated by the convolutional backbone.

\section{Method}
\label{sec:method}

\subsection{Overview}
An overview of ADHMR is presented in Figure~\ref{fig:overview}.
Given an input image $I$, we aim to reconstruct the 3D human mesh in a parameterized way, which is to predict the pose parameters ${\theta} \in \mathbb{R}^{24 \times 3}$ and shape parameters ${\beta} \in \mathbb{R}^{10}$ of the predefined SMPL model~\cite{pavlakos2019smplxehf}. We formulate this problem as a generation process conditioned on the input image to tackle the inherent reconstruction ambiguity.

We begin by training a diffusion-based HMR base model (Sec.~\ref{sec:scorehypo}).
Next, we construct a synthetic human preference dataset (Sec.~\ref{sec:dpodataset}), where candidate human mesh predictions are generated by the base model and then paired based on scores provided by the assessment model HMR-Scorer (Sec.~\ref{sec:scorer}).
To distill knowledge from this synthetic preference dataset, we propose a preference optimization framework ADHMR that finetunes the base model to preferentially generate winner predictions over losers (Sec.~\ref{sec:dpo}).
Furthermore, thanks to the strong capacity of HMR-Scorer to assess mesh predictions, we filter training data to enhance several popular HMR models (Sec.~\ref{sec:dataclean}).

\subsection{HMR-Scorer}
\label{sec:scorer}
Given a set of predictions $\left\{\mathbf{P}_m = (\theta, \beta, \Pi)_m \right\}_{m=0}^M$ for an input image $I$, HMR-Scorer aims to assign an estimated quality score $\left\{s_m \in \mathbb{R}\right\}_{i=0}^M$ to each prediction. $M$ represents the number of predictions, $\Pi$ is the predicted camera parameters. Higher scores should be assigned to predictions with higher quality and better aligned with the image.

\noindent
\textbf{Model architecture.} 
We first introduce the input features into HMR-Scorer.
Instead of directly encoding pose parameters, which are prone to ambiguities in representing joint positions and deficient in spatial context, we leverage UVD coordinates as the input to HMR-Scorer. This provides a unified and consistent representation of the 3D human skeleton and preserves the geometric relationships of the skeleton.
Specifically, using the camera model, we first project human body keypoints to the input image space to get their UVD coordinates $\mathbf{J}_{uvd} \in \mathbb{R}^{N \times 3}$, where $N=29$ is the number of keypoints.
We use a multilayer perceptron (MLP) to map $\mathbf{J}_{uvd}$ to a high-dimensional feature vector $\mathbf{F}^J \in \mathbb{R}^{C^l \times N}$.

We utilize multi-scale image features as the image condition, denoted as $\mathbf{c}:=\left\{\mathbf{F}^g, \mathbf{F}^l\right\}$.
The input image $I$ is initially divided into fixed-size patches through a patch embedding mechanism, producing a sequence of image tokens. These tokens are subsequently processed using a ViT-Base model~\cite{dosovitskiy2020vit} to generate a series of global image feature tokens, denoted as $\mathbf{F}^g \in \mathbb{R}^{C^g \times H^g \times W^g}$. The global image feature tokens are then passed through a convolutional network to derive the low-channel global features, represented as $\mathbf{F}^g \in \mathbb{R}^{C^l \times H^g \times W^g}$.
A de-convolution head is deployed on the global feature $\mathbf{F}^{g}$ to obtain the high-resolution local feature map $\mathbf{F}^l \in \mathbb{R}^{C^l \times H^l \times W^l}$. $C^*$ and $H^* \times W^*$ denote the feature channel and size, respectively. 
We sample the local feature $\mathbf{F}^l$ according to the re-projected 2D joint positions $\mathbf{J}_{uv}$ and obtain pixel-aligned local image features $\mathbf{F}^L \in \mathbb{R}^{C^l \times N}$ for each joint.

The concatenated features of $\mathbf{F}^J$ and $\mathbf{F}^L$ are subsequently fed into a transformer-encoder-based network comprising $B$ fundamental blocks. Each block integrates a multi-head self-attention (MHSA) mechanism, a cross-attention (CA) unit, and a feed-forward network (FFN). Within the CA unit, the global image feature $\mathbf{F}^g$ serves as the key and value features, while the query feature is derived from the output of the preceding MHSA unit. Through the cross-attention mechanism, the geometric information from the human mesh predictions is effectively aligned with image features, ensuring a coherent integration of structural and visual cues.
Finally, a decoder network, implemented as an MLP, is employed to estimate the score $s$.

\noindent
\textbf{Training.}
We construct a training dataset comprising human mesh predictions for corresponding images and their quality labels to train the HMR-Scorer.
Specifically, predictions are generated by adding joint-wise Gaussian noise to the ground truth SMPL pose to simulate rotational errors, with magnitudes empirically determined. The reconstruction quality labels are annotated using standard HMR metrics, including PVE $Q^{pve}$, MPJPE $Q^{mpjpe}$, PA-MPJPE $Q^{pajpe}$, and PA-PVE $Q^{papve}$. Details of these metrics are provided in Sec.~\ref{subsec:hmrmetrcs}.
To accurately capture subtle quality differences, the training process is designed to learn relative quality preferences among predictions. Inspired by RankNet~\cite{burges2005ranknet}, we utilize a probabilistic ranking cost function:
\begin{equation}
    \begin{aligned}
        \mathcal{L}_{mn}\left(s_{m n}, y_{m n}\right):= & -y_{m n} \log s_{m n} \\
        & -\left(1-y_{m n}\right) \log \left(1-s_{m n}\right),
    \end{aligned}
\end{equation}
where $s_{m n} = \mathrm{Sigmoid}\left( s_m - s_n \right)$ is the relative quality difference probability between predictions $\mathbf{P}_m$ and $\mathbf{P}_n$, and $y_{mn}$ is the ground truth quality label representing the quality difference between predictions based on each of the above-mentioned four HMR metrics. For instance, for the PVE benchmark, the label is represented as $y_{mn}^{pve}$ ($y_{mn}^{pve} = 1$ if $Q^{pve}_m < Q^{pve}_n$, and $0$ otherwise). The overall training loss for HMR-Scorer is defined as follows:
\begin{equation}
    \begin{aligned}
        & \mathcal{L}^{\text {HMR-Scorer}}  =  \\
        & \sum_{m,n=0; n \neq m}^M \left(\mathcal{L}_{mn}^{pve}+\mathcal{L}_{mn}^{papve}+\mathcal{L}_{mn}^{pajpe}+\mathcal{L}_{mn}^{mpjpe} \right).
    \end{aligned}
\end{equation}

\begin{table*}
\centering
    \resizebox{0.96\linewidth}{!}
    {
    \begin{tabular}{l|cc|cc|cc|cc|cc|cc}
    \toprule
    \multirow{3}{*}{}  & \multicolumn{6}{c|}{ GTA-Human } & \multicolumn{6}{c}{ DNA-Rendering }   \\
     \cmidrule{2-13}
     & \multicolumn{2}{c|}{ PVE } & \multicolumn{2}{c|}{ MPJPE } & \multicolumn{2}{c|}{ PA-MPJPE } & \multicolumn{2}{c|}{ PVE } & \multicolumn{2}{c|}{ MPJPE} & \multicolumn{2}{c}{ PA-MPJPE } \\
    \cmidrule{2-13}  & PLCC $\uparrow$ & SRCC $\uparrow$  & PLCC $\uparrow$ & SRCC $\uparrow$ & PLCC $\uparrow$ & SRCC $\uparrow$ & PLCC $\uparrow$ & SRCC $\uparrow$ & PLCC $\uparrow$ & SRCC $\uparrow$ & PLCC $\uparrow$ & SRCC $\uparrow$  \\
    \midrule 
    ScoreNet~\cite{xu2024scorehypo} & 0.52 & 0.49 & 0.52 & 0.50 & 0.47 & 0.43 & 0.55 & 0.51 & 0.55 & 0.50 & 0.50 & 0.46  \\
    HMR-Scorer-P & 0.30 & 0.28 & 0.28 & 0.26 & 0.29 & 0.25 & 0.34 & 0.31 & 0.32 & 0.29 & 0.34 & 0.27  \\
    HMR-Scorer-2D & 0.59 & 0.58 & 0.59 & 0.58 & 0.50 & 0.49 & 0.62 & 0.60 & 0.63 & 0.61 & 0.56 & 0.53  \\
    \textbf{HMR-Scorer (Ours)} & \textbf{0.63} & \textbf{0.62} & \textbf{0.63} & \textbf{0.62} & \textbf{0.57} & \textbf{0.54} & \textbf{0.66} & \textbf{0.64} & \textbf{0.66} & \textbf{0.64} & \textbf{0.62} & \textbf{0.58} \\
    \bottomrule
    \end{tabular}
    }
    \vspace{-5pt}
    \caption{Score prediction results on the GTA-Human~\cite{cai2024gtahuman} and DNA-Rendering~\cite{cheng2023dnarender} dataset. We report the PLCC and SRCC between the predicted scores and the PVE, MPJPE, and PA-PVE ground truth errors, respectively.
    }
    \vspace{-10pt}
    \label{tab:scorer}
\end{table*}

\subsection{HMR Preference Dataset Construction}
\label{sec:dpodataset}
We leverage preference-based optimization rather than traditional supervised training to finetune the base HMR model.
However, traditional preference optimization methods require human preference datasets labeled by human annotators, which are currently unavailable for this field.

To this end, we propose to use HMR-Scorer to synthesize an HMR preference dataset $\mathcal{D}=\left\{\left(I, \mathbf{x}_0^w, \mathbf{x}_0^l\right)\right\}$, where each sample contains the input image $I$ and a pair of human mesh predictions generated from the HMR base model $\boldsymbol{\epsilon}_\text{ref}$.
Specifically, given a set of predictions $\left\{\mathbf{P}_m\right\}_{m=0}^M$, HMR-Scorer assigns scores $\left\{s_m \in \mathbb{R}\right\}_{m=0}^M$ for these predictions. According to the scores, these predictions undergo ordinal arrangement, which emulates human preference in ranking human mesh reconstructions, from highest to lowest fidelity.
This hierarchical organization facilitates the extraction of paired samples $\left(\mathbf{P}^w, \mathbf{P}^l \right)$, denoting superior (winner) and inferior (loser) predictions respectively, where their associated scores satisfy the preference relation $\left(\mathbf{P}^w \succ \mathbf{P}^l \mid I \right)$. The pairing process involves stochastic selection of winners from the top $K$ highest-scoring predictions, coupled with losers from the $K$ lowest-scoring predictions, generating $K$ distinct pairs per image.
For studio-captured datasets with precise human mesh annotations, the prediction quality ordering is directly determined by computing the reconstruction error against ground truth labels.

\subsection{ADHMR}
\label{sec:dpo}
ADHMR aligns the base model $\boldsymbol{\epsilon}_\text{ref}$ with the constructed preference dataset $\mathcal{D}=\left\{\left(I, \mathbf{x}_0^w, \mathbf{x}_0^l\right)\right\}$ to produce superior predictions.
The aligned model $\boldsymbol{\epsilon}_\theta$ is initialized using the parameters of the base model $\boldsymbol{\epsilon}_\text{ref}$, while keeping the base model's parameters frozen throughout the training process. 
The proposed optimization framework extends the direct preference optimization (DPO) method. The principle of DPO lies in its direct optimization of a conditional distribution $\boldsymbol{\epsilon}_\theta\left(\mathbf{x}_0 \mid \mathbf{c}\right)$, contrasting with RLHF's approach of optimizing a reward model $r\left(\mathbf{c}, \mathbf{x}_0\right)$, while simultaneously constraining the KL-divergence from a reference distribution $\boldsymbol{\epsilon}_\text{ref}$:
\begin{equation}
    \begin{aligned}
        & \max _{\boldsymbol{\epsilon}_\theta} \mathbb{E}_{\mathbf{c} \sim \mathcal{D}_c, \mathbf{x}_0 \sim \boldsymbol{\epsilon}_\theta \left(\mathbf{x}_0 \mid \mathbf{c}\right)} \\
         & \left[r\left(c, \mathbf{x}_0\right)\right] 
         - \beta_{\operatorname{KL}}\left[\boldsymbol{\epsilon}_\theta\left(\mathbf{x}_0 \mid \mathbf{c}\right) \| \boldsymbol{\epsilon}_{\text {ref }}\left(\mathbf{x}_0 \mid \mathbf{c}\right)\right].
    \end{aligned}
\end{equation}
Following \cite{wallace2024diffusiondpo}, a significant challenge in applying DPO to diffusion models is the intractability of the parameterized distribution $\boldsymbol{\epsilon}_\theta\left(\mathbf{x}_0 \mid \mathbf{c}\right)$, which stems from the necessity to marginalize over all possible diffusion trajectories $\left(\mathbf{x}_1, \ldots, \mathbf{x}_T\right)$ that culminate in $\mathbf{x}_0$. Through some mathematical techniques, this challenge is addressed by formulating an objective function that operates on the complete denoising trajectory $\mathbf{x}_{0: T}$:
\begin{equation}
    \begin{gathered}
        \mathcal{L}^\text{DDPO}(\theta)=-\mathbb{E}_{(\mathbf{x}_0^w, \mathbf{x}_0^l) \sim \mathcal{D}, t \sim \mathcal{U}(0, T), \mathbf{x}_t^w \sim q(\mathbf{x}_t^w \mid \mathbf{x}_0^w), \mathbf{x}_t^l \sim q(\mathbf{x}_t^l \mid \mathbf{x}_0^l)} \\
        \log \sigma(-\beta T \omega\left(\lambda_t\right)( \\
        \left\|\boldsymbol{\epsilon}^w-\boldsymbol{\epsilon}_\theta\left(\mathbf{x}_t^w, t\right)\right\|_2^2-\left\|\boldsymbol{\epsilon}^w-\boldsymbol{\epsilon}_{\mathrm{ref}}\left(\mathbf{x}_t^w, t\right)\right\|_2^2 \\
        \left.\left.-\left(\left\|\boldsymbol{\epsilon}^l-\boldsymbol{\epsilon}_\theta\left(\mathbf{x}_t^l, t\right)\right\|_2^2-\left\|\boldsymbol{\epsilon}^l-\boldsymbol{\epsilon}_{\mathrm{ref}}\left(\mathbf{x}_t^l, t\right)\right\|_2^2\right)\right)\right)
    \end{gathered}
\end{equation}
where $\mathbf{x}_t^*=\alpha_t \mathbf{x}_0^*+\sigma_t \boldsymbol{\epsilon}^*$ is drawn from $q\left(\mathbf{x}_t^* \mid \mathbf{x}_0^*\right)$ with $\boldsymbol{\epsilon}^* \sim \mathcal{N}(0, \mathbf{I})$. Here, $\lambda_t=\alpha_t^2 / \sigma_t^2$ denotes the signal-to-noise ratio, $\omega\left(\lambda_t\right)$ is a weighting function, and the constant $T$ is factored into $\beta$.

During training, the model improves by comparing points along the diffusion trajectory with examples from the synthetic preference dataset. This helps the model better denoise winner mesh predictions compared to losers, as evaluated by HMR-Scorer.
Hence, this methodology guides the model to generate human mesh predictions that not only align closely with the input image but also adhere to a realistic distribution of human poses. 
By exclusively finetuning the denoiser component within the latent space, the approach achieves more generalized and well-aligned results without overfitting to noisy labels, especially for in-the-wild datasets.

\subsection{Data Cleaning}
\label{sec:dataclean}
To further evaluate the efficacy of our trained HMR-Scorer, we propose to apply it to the training data cleaning process, aiming to determine whether the proposed scorer can effectively identify noisy data. While many indoor datasets have ground truth labels, in-the-wild datasets often rely on noisy pseudo labels, which hinders model training and generalization. Therefore, we use HMR-Scorer to remove low-quality samples, ensuring a reliable training dataset.

The data cleaning process begins with score computation, where a quality score \( s_i \in [0, 1] \) is assigned to each sample \( (I_i, \hat{\theta}_i, \hat{\beta}_i, \hat{\Pi}_i) \) in the dataset \( \mathcal{X} \) using HMR-Scorer. 
The score evaluates the alignment of predictions with the input image and the plausibility of the model parameters. Next, a threshold \( \tau \) is applied to filter out low-quality samples, resulting in the cleaned dataset \( \mathcal{X}_{\text{clean}} = \{(I_i, \hat{\theta}_i, \hat{\beta}_i, \hat{\Pi}_i) \mid s_i \geq \tau\} \). Finally, only high-confidence pseudo-labels are retained for model training. 
\begin{table*}
    \centering
    \resizebox{0.85 \linewidth}{!}
    {
    \begin{tabular}{l|c|ccc|ccc}
    \toprule
    \multirow{2}{*}{ Method } & \multirow{2}{*}{$M$} & \multicolumn{3}{c|}{ 3DPW } & \multicolumn{3}{c}{ Human3.6M }   \\
     \cline{3-8}
     & & { PVE $\downarrow$ } & { MPJPE $\downarrow$ } & { PA-MPJPE $\downarrow$ } & { PVE $\downarrow$ } & { MPJPE $\downarrow$ } & { PA-MPJPE $\downarrow$ } \\
    \midrule
    HMR~\cite{kanazawa2018hmr}         & - & 152.7 & 130.0 & 81.3 & 96.1 & 88.0 & 56.8 \\
    HybrIK~\cite{li2021hybrik}       & - & 86.5  & 74.1  & 45.0 & 65.7 & 54.4 & 34.5 \\
    PyMaf~\cite{zhang2021pymaf}         & -  & 110.1 & 92.8 & 58.9  & -    & 57.7 & 40.5 \\
    POTTER~\cite{zheng2023potter}         & -  & 87.4  & 75.0 & 44.8 & -    & 56.5 & 35.1  \\
    ImpHMR~\cite{cho2023imphmr}         & -  & 87.1  & 74.3 & 45.4 & -    & -    & -  \\
    Zolly~\cite{wang2023zolly}            & -  & 76.3  & 65.0 & 39.8 & - & 49.4 & 32.3 \\
    HMR 2.0a~\cite{goel2023hmr2.0} & - & -    & 70.0 & 44.5 & -  & 44.8 & 33.6 \\
    HMR 2.0b~\cite{goel2023hmr2.0} & - & -    & 81.3 & 54.3 & -  & 50.0 & 32.4 \\
    ScoreHMR~\cite{stathopoulos2024scorehmr} & - & - & 76.8 & 51.1 & -  & - &  \\
    \midrule
    \multirow{2}{*}{Biggs \textit{et al.}~\cite{biggs20203dmultibodies}} & 10  & -    & 79.4 & 56.6 & -    & 59.2 & 42.2 \\
    & 25  & -    & 75.8 & 55.6 & -    & 58.2 & 42.2 \\
    Sengupta \textit{et al.}~\cite{sengupta2021hierarchical} & 25    & -    & 75.1 & 47.0    & -    & -    & -   \\
    ProHMR~\cite{kolotouros2021prohmr} & 25 & -    & -    & 52.4 & -    & -    & 36.8 \\
    HuManiFlow ~\cite{sengupta2023humaniflow} & 100 & -    & 65.1 & 39.9  & -    & -    & -  \\
    HMDiff ~\cite{foo2023hmdiff} & 25   & 82.4 & 72.7 & 44.5  & -    & 49.3 & 32.4 \\
    \midrule
    \multirow{3}{*}{HypoNet (Base Model) } & 10 & 79.8 & 68.5 & 41.0 & 52.5 & 42.4 & 29.0  \\
    & 100 & 73.4 & 63.0 & 37.6 & 47.5 & 38.4 & 26.0  \\
    & 200 & 71.9 & 61.8 & 36.1 & 46.4 & 37.4 & 25.3  \\
     
    \midrule 
    \multirow{3}{*}{ \textbf{ADHMR} } & 10 & 73.8 & 64.2 & 38.3 & 52.1 & 41.8 & 28.4 \\
     & 100 & {65.4} & {57.2} & {33.5} & 45.9 & 36.9 & 24.8 \\
     & 200 & 63.5 & {55.7} & {32.5} & {44.6} & \textbf{35.8} & {24.1} \\
    \midrule 
    \multirow{3}{*}{ \textbf{ADHMR (ITW)} } & 10 & 71.3 & 61.3 & 37.1 & 52.2 & 41.9 & 28.3 \\
     & 100 & 62.6 & 54.2 & 32.0 & 45.9 & 37.0 & 24.8 \\
     & 200 & \textbf{60.5} & \textbf{52.6} & \textbf{30.8} & \textbf{44.6} & {35.9} & \textbf{24.0} \\
    \bottomrule
    \end{tabular}
    }
    \vspace{-5pt}
    \caption{Comparison with state-of-the-arts on the 3DPW~\cite{3dpw} and Human3.6M~\cite{human36m} dataset. 
     $M$ is the number of predictions in probabilistic methods.
     ADHMR is finetuned on the target benchmark dataset, while ADHMR (ITW) is further finetuned on the preference dataset constructed from an in-the-wild dataset.
    }
    \vspace{-10pt}
    \label{tab:dpo}
\end{table*}

\section{Experiments}
\label{sec:exp}

\subsection{Setup}
\label{sec:expsetup}

\noindent
\textbf{Training.}
HMR-Scorer is trained on five datasets, including HI4D~\cite{yin2023hi4d}, BEDLAM~\cite{black2023bedlam}, DNA-Rendering~\cite{cheng2023dnarender}, GTA-Human~\cite{cai2024gtahuman}, and SPEC~\cite{kocabas2021spec}. These datasets contain accurate 3D annotations for human pose, which plays an important role in training an effective scorer. The HMR base model is the current state-of-the-art probabilistic HMR method: HypoNet from \cite{xu2024scorehypo}.

\noindent
\textbf{Evaluation metrics.}
We use four standard human mesh reconstruction metrics: PVE, MPJPE, PA-PVE, and PA-MPJPE, as detailed in Sec.~\ref{subsec:hmrmetrcs}.
To evaluate the scorer, we follow score prediction assessment~\cite{zhai2020iqasurvey} to employ two standard metrics: the Pearson linear correlation coefficient (PLCC) and Spearman rank correlation coefficient (SRCC). These correlation coefficients quantify the alignment between the predicted scores and ground truth reconstruction error.

\subsection{HMR-Scorer Evaluation}
\noindent
\textbf{Test benchmark.}
Since there are no existing datasets for score prediction of HMR models, we construct a test set from a synthetic dataset GTA-Human~\cite{cai2024gtahuman}, which is produced with rendering engines (e.g., Unreal Engine) and contains accurate 3D annotations.
We also use DNA-Rendering~\cite{cheng2023dnarender}, a large-scale multi-view studio-based dataset with an ultra-high resolution, to construct the other test set to show the capacity for common studio-based scenes.
We adopt the original test set of the two selected datasets. Then we perturb the ground truth pose labels with random gaussian noise to simulate predictions of HMR models. We save the corresponding reconstruction errors for the noised predictions. 
We then measure the PLCC and SRCC between the ground truth reconstruction metrics and the predicted scores.

\noindent
\textbf{Results.}
Table~\ref{tab:scorer} presents the comparison results for score prediction. 
We observe that our scorer outperforms all baseline methods in both PLCC and SRCC metrics, showcasing its efficacy.
For comparison, we modify the reward model in ScoreHypo as a baseline.
We also study two ablations of HMR-Scorer: HMR-Scorer-P accepts SMPL(-X) pose rotation vectors as input, and HMR-Scorer-2D accepts 2D joint positions (without joint depth) as input.
Results show that HMR-Scorer outperforms the baselines in aligning the scores with the real reconstruction errors, which underscores its effectiveness indicating human mesh prediction quality.

\begin{table}
    \centering
    \resizebox{0.92\linewidth}{!}
    {
    \begin{tabular}{l|ccc}
    \toprule
    \multirow{2}{*}{ Method } & \multicolumn{3}{c}{ 3DPW }   \\
     \cline{2-4}
     & { PVE $\downarrow$ } & { MPJPE $\downarrow$ } & { PA-MPJPE $\downarrow$ } \\
    \midrule
    HypoNet (Base Model) & 73.4 & 63.0 & 37.6   \\
     
    \midrule 
    \multicolumn{4}{c}{ \textit{(a) Finetune on target benchmark dataset} } \\
    \midrule
    Supervised finetuning & 68.0 & 59.4 & 35.9  \\
    {ADHMR} & {\textbf{65.4}} & {\textbf{57.2}} & {\textbf{33.5}}  \\
     \midrule
    \multicolumn{4}{c}{ \textit{(b) Finetune on in-the-wild dataset} } \\
    \midrule
    Supervised finetuning & 70.2 & 61.3 & 36.5  \\
    ADHMR & \textbf{62.6} & \textbf{54.2} & \textbf{32.0} \\
    \bottomrule
    \end{tabular}
    }
    \caption{Ablation of preference finetuning on 3DPW~\cite{3dpw} dataset. $M=100$ for all models.
    }
    \vspace{-20pt}
    \label{tab:ablation1}
\end{table}

\begin{table}
    \centering
    \resizebox{1.\linewidth}{!}
    {
    \begin{tabular}{l|ccc}
    \toprule
    \multirow{2}{*}{ Method } & \multicolumn{3}{c}{ 3DPW } \\
     \cline{2-4}
     & { PVE $\downarrow$ } & { MPJPE $\downarrow$ } & { PA-MPJPE $\downarrow$ } \\
    \midrule
    HypoNet (Base Model) & 71.9 & 61.8 & 36.1  \\     
    Supervised finetuning & 69.9 & 59.7 & 35.2 \\
    ADHMR (ITW) & \textbf{60.5} & \textbf{52.6} & \textbf{30.8} \\
    \bottomrule
    \end{tabular}
    }
    \vspace{-5pt}
    \caption{Ablation of extra training data on 3DPW~\cite{3dpw} dataset. We use multiple training datasets of the scorer to perform supervised finetuning. $M=200$ for all models.
    }
    \vspace{-10pt}
    \label{tab:ablation2}
\end{table}

\begin{figure*}
    \centering
    \includegraphics[width=0.925\linewidth]{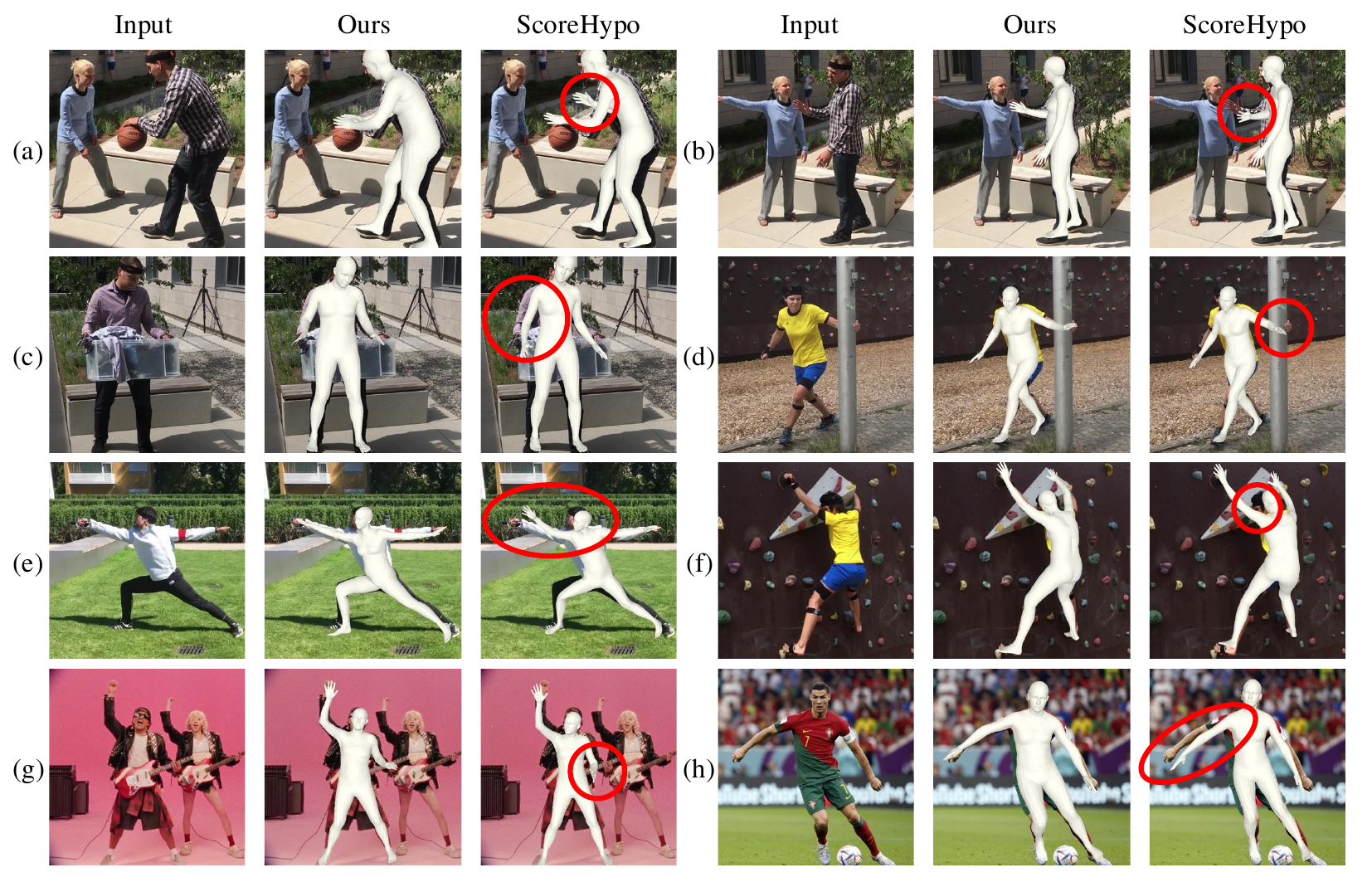}
    \vspace{-10pt}
    \caption{Qualitative comparison of the state-of-the-art probabilistic model ScoreHypo~\cite{xu2024scorehypo} and our ADHMR. Our framework significantly improves image alignment and in-the-wild robustness. (a) $\sim$ (f) are from the 3DPW~\cite{3dpw} dataset, and (g) $\sim$ (h) are challenging in-the-wild images.}
\label{fig:qualitative}
\end{figure*}

\begin{table*}
    \centering
    \resizebox{0.958\linewidth}{!}
    {
    \begin{tabular}{l|cc|cc|ccc}
    \toprule
    \multirow{3}{*}{Method}  & \multicolumn{2}{c|}{ 3DPW } & \multicolumn{2}{c|}{ Human3.6M } & \multicolumn{3}{c}{ EHF }   \\
    \cmidrule{2-8}
     & { MPJPE $\downarrow$} & { PA-MPJPE $\downarrow$} & { MPJPE $\downarrow$} & { PA-MPJPE $\downarrow$} & { PA-PVE $\downarrow$} & { PVE $\downarrow$} & { PA-MPJPE $\downarrow$}\\
    \midrule 
    Hand4Whole ~\cite{moon2022h4w} & 115.2 & 75.4 & 78.8 & 57.7 & 57.8 & 89.2 & 70.2  \\
    + data cleaning & \textbf{112.3} & \textbf{73.7} & \textbf{77.9} & \textbf{57.0} & \textbf{56.2} & \textbf{88.8} & \textbf{69.8} \\
    \midrule
    OSX (Base) \cite{lin2023osx} & 100.4 & 66.4 & 69.5 & 48.9 & 54.7 & 86.6 & 63.7 \\
    + data cleaning & \textbf{99.4} & \textbf{65.2} & \textbf{65.7} & \textbf{46.4} & \textbf{53.1} & \textbf{84.0} & \textbf{62.4} \\
    \midrule
    SMPLer-X-Base \cite{cai2024smplerx} & 99.5 & 64.2 & 59.8 & 45.8 & \textbf{51.0} & 82.4 & 59.7 \\
    + data cleaning  & \textbf{97.9} & \textbf{62.9} & \textbf{57.5} & \textbf{43.8} & 51.4 & \textbf{78.6} & \textbf{58.6} \\
    \bottomrule
    \end{tabular}
    }
    \caption{Quantitative comparisons between several state-of-the-art methods with and without data cleaning on the 3DPW~\cite{3dpw}, Human3.6M~\cite{human36m} and EHF~\cite{pavlakos2019smplxehf} dataset. 
    All methods are trained on four commonly used datasets. After cleaning the training data, these models achieve higher accuracy even when trained on a smaller subset.}
    \vspace{-4pt}
    \label{tab:clean}
\end{table*}

\subsection{ADHMR Evaluation}
\noindent
\textbf{Comparisons with state-of-the-art methods.}
In Table~\ref{tab:dpo}, we compare the accuracy of the ADHMR with state-of-the-art methods on two widely used benchmark datasets Human3.6M~\cite{human36m} and 3DPW~\cite{3dpw}. 3DPW is an in-the-wild dataset.
We show results of both deterministic and probabilistic methods.
Following the conventions of standard probabilistic approaches~\cite{xu2024scorehypo, biggs2020multibodies}, we generate multiple estimates and report the min-MPJPE and min-PVE of the $M$ predictions.
When finetuned directly on the target benchmark dataset, ADHMR achieves further enhancements, showcasing its strong ability to adapt to and balance domain-specific distribution. 
To evaluate the effectiveness under the in-the-wild setting, ADHMR (ITW) is further finetuned on InstaVariety dataset~\cite{kanazawa2019instavariety}, which contains various in-the-wild images annotated with noisy pseudo-labels collected from Instagram. Please note that we do not use the original 3D labels but use the HMR-Scorer to construct a preference dataset.
Notably, we observe that our finetuned model achieves better performance using $M=10$ predictions than the base model using $M=200$ predictions on the in-the-wild benchmark 3DPW, showcasing that our finetuning pipeline greatly enhances the generalizability to in-the-wild datasets and its efficiency.
Moreover, ADHMR consistently outperforms existing state-of-the-art methods by a substantial margin.

\noindent
\textbf{Qualitative results.}
Fig.~\ref{fig:qualitative} shows qualitative comparisons between ADHMR and the previous state-of-the-art probabilistic method ScoreHypo.
We show randomly selected results of ScoreHypo and ADHMR on 3DPW and internet images with $M=10$ candidate predictions.
We can see that the finetuned model can produce more accurate results for body pose under challenging cases, such as dense human-environment interactions. The base model, however, cannot achieve good image-mesh alignment.
For example, in the (a) instance, our method provides more reasonable poses for the occluded right arm. In the (c) instance, ScoreHypo gives erroneous prediction for the person's arms, while ours gives a more accurate body pose prediction, proving the efficacy of the proposed finetuning pipeline. In the (g) instance, ScoreHypo produces inaccurate elbow poses, while our prediction fits the input image better. Results show that ADHMR is more robust for challenging internet images than ScoreHypo.
Please zoom in to observe our improvement over the base model.

\noindent
\textbf{Ablation of preference finetuning.}
As shown in Table~\ref{tab:ablation1}, we conduct an ablation study on different finetuning methods to demonstrate the advantages of ADHMR over traditional supervised finetuning. We construct two baselines where the base model is finetuned using the ground truth labels in the datasets.
We first finetune the base model on the training sets of the two target benchmarks (3DPW and Human3.6M). We also finetune on the pseudo labels of InstaVariety to simulate training on noisy pseudo-labeled in-the-wild data.
Results show that ADHMR consistently outperforms traditional supervised finetuning in both settings. 
When finetuned directly on the target benchmark dataset, our method achieves further enhancements than supervised finetuning, showcasing its strong ability to adapt to and balance domain-specific distribution. 
In the meantime, supervised finetuning may overfit one training dataset (3DPW) and the performance on the other test benchmark (Human3.6M) could be corrupted.
On the in-the-wild dataset, ADHMR demonstrates robustness under noisy pseudo-label conditions, while supervised finetuning on the noisy labels could overfit the training dataset and harm its performance on 3DPW benchmark.

\noindent
\textbf{Ablation of extra training datasets.}
In Table~\ref{tab:ablation2}, we aim to confirm that the improvements achieved by ADHMR are contributed to its inherent optimization strategy, rather than including extra information from other datasets used during scorer training.
So we additionally finetune the base model on the scorer's training sets.
Supervised finetuning on scorer training datasets yields insignificant performance gains, confirming that the improvements are not primarily due to external dataset information. 
This suggests that simply exposing the base model to more data does not guarantee better performance.
In contrast, ADHMR benefits from the ability of HMR-Scorer to implicitly guide the base model towards generating high-quality predictions.

\subsection{Data Cleaning Results}
Current HMR models are trained using quite different datasets. For a fair and comprehensive comparison, we selected several state-of-the-art HMR methods and retrained them using four commonly used datasets: MSCOCO~\cite{lin2014mscoco}, MPII~\cite{andriluka2014mpii}, Human3.6M~\cite{human36m}, and MPI-INF-3DHP~\cite{mehta20173dhp}.
We compare training the models on the full training sets of these datasets and training on the filtered datasets obtained after applying data cleaning with the HMR-Scorer. We set the filter threshold $\tau=0.6$.

Quantitative results are in Table~\ref{tab:clean}. The results demonstrate that models trained on the cleaned datasets achieve better performance despite using less training data. This improvement highlights the utility of our method in filtering out low-quality training samples, thereby enabling the models to focus on higher-quality data for learning. 
This finding provides a new perspective for improving large HMR models by strategically cleaning training data. By leveraging the HMR-Scorer to curate datasets, we can achieve higher-quality model training with even less data, making it a valuable tool for effortless performance gain.

\section{Conclusion}
\label{sec:conclusion}

In this work, we propose ADHMR, the first framework for aligning diffusion-based HMR models with direct preference optimization. 
We leverage a trained HMR-Scorer to synthesize a preference dataset automatically without the need for manual annotation.
This dataset is then used to align the diffusion-based HMR model through direct preference optimization. 
Additionally, HMR-Scorer improves the performance of several state-of-the-art HMR models by filtering out low-quality training data.
Extensive experiments validate the effectiveness of ADHMR.
We believe that our work will pave the way for future advancements in alignment techniques for human mesh recovery.

\section*{Acknowledgements}

This study is supported under the RIE2020 Industry Alignment Fund – Industry Collaboration Projects (IAF-ICP) Funding Initiative, as well as cash and in-kind contribution from the industry partner(s). This research is also supported by A*STAR under its MTC Programmatic Funds (Grant No. M23L7b0021) and the MoE AcRF Tier 1 grant (RG14/22).

\section*{Impact Statement}

This paper presents work whose goal is to advance the field of Machine Learning. There are many potential societal consequences of our work, none which we feel must be specifically highlighted here.


\bibliography{main}
\bibliographystyle{icml2025}




\end{document}